\documentclass{article}

\PassOptionsToPackage{numbers, compress}{natbib}
\usepackage[final]{nips_2018}

\usepackage[utf8]{inputenc} % allow utf-8 input
\usepackage[T1]{fontenc}    % use 8-bit T1 fonts
\usepackage{hyperref}       % hyperlinks
\usepackage{url}            % simple URL typesetting
\usepackage{booktabs}       % professional-quality tables
\usepackage{amsfonts}       % blackboard math symbols
\usepackage{nicefrac}       % compact symbols for 1/2, etc.
\usepackage{microtype}      % microtypography
\usepackage[pdftex]{graphicx}
\title{Towards Neural Machine Translation for African Languages}

\author{
  Jade Z. Abbott \\
  Retro Rabbit\\
  \texttt{ja@retrorabbit.co.za} \\
   \And
  Laura Martinus \\
  Human Language Technologies, CSIR\\
  \texttt{lmartinus@csir.co.za} \\
}

\begin{document}

\maketitle

\begin{abstract}
   Given that South African education is in crisis, strategies for improvement and sustainability of high-quality, up-to-date education must be explored. In the migration of education online, inclusion of machine translation for low-resourced local languages becomes necessary. This paper aims to spur the use of current neural machine translation (NMT) techniques for low-resourced local languages. The paper demonstrates state-of-the-art performance on English-to-Setswana translation using the Autshumato dataset. The use of the Transformer architecture beat previous techniques by 5.33 BLEU points. This demonstrates the promise of using current NMT techniques for African languages.
\end{abstract}

\section{Introduction}

Given that South Africa's education system is in crisis \cite{spaull2013south}, strategies for sustainability within education must be explored. One suggested strategy to improve youth education would be to augment the learning process with online content \cite{parikh2002utilizing}. %One suggested strategy is... Or something. The two sentences aren't linked

The internet comprises of 53.5\% English content, while the other 10 official South African languages comprise of less than 0.1\% of the languages spoken on the internet \cite{w3tech2018usage}. According to the 2011 South African census, only 9.8\% of South Africans speak English as a primary language \cite{stats2012census}. Similar statistics exist for many other African countries \cite{chepkemoi_2017,nationmaster,muaka2011language}. 
%TODO: insert citations 

For the South African youth to benefit from the massive amount of educational online content, the translation of online resources into the many low-resourced African languages is sorely needed. Unfortunately, machine translation of low-resourced languages has proven difficult with both conventional statistical machine translation (SMT) and the more recent neural machine translation (NMT) methods \cite{peter2016rwth, gu2018universal,ronald2006statistical,wilken2012developing}.

The convolutional sequence-to-sequence (ConvS2S) architecture improved translation results on multiple languages, including low-resourced languages \cite{gehring2017convolutional}. Additionally, by pre-training the Transformer architecture on many languages and then specialising to a single language, Gu \textit{et al.} were able to improve performance on low-resourced languages \cite{gu2018meta}.

This paper aims to serve as the initial work towards using modern neural machine translation (NMT) techniques to improve machine translation on African languages, and invigorate future research into using such techniques. NMT techniques are often overlooked in favour of conventional phrase-based translation systems. This is due to vanilla NMT's reputation for comparatively under-performing on low-resource languages \cite{gu2018meta}. This paper shows the performance of training convolutional sequence-to-sequence learning \cite{gehring2017convolutional} and the Transformer architecture on the Autshumato English-Setswana Parallel Corpora \cite{cmckellar2018autshumato}.

\section{Related Work}

Kato \textit{et al.} used statistical phrase-based translation, based on Moses, in order to perform English-to-Setswana translation \cite{ronald2006statistical}. They achieve a BLEU score of 32.71 on a dataset that is not publically-available and so was excluded from the comparison. Wilken \textit{et al.} used a similar technique as \cite{ronald2006statistical}, but focused on linguistically-motivated pre- and post-processing of the corpus in order to improve translation with phrase-based techniques \cite{wilken2012developing}. Wilken \textit{et al.} was trained on the same Autshumato dataset used in this paper, and also used an additional monolongual dataset for language modelling.

\section{Methodology}

Section~\ref{dataset} describes the parallel dataset used, while the selected models and their hyperparameters for training are described in Section~\ref{models}.

\subsection{Dataset}
\label{dataset}
The publically-available Autshumato English-Setswana Parallel Corpora is an aligned corpus of South African governmental data which was created for the use in machine translation systems. The dataset consists of three smaller parallel corpora obtained from different sources which were combined to form a single corpus. The combined corpus was sharded into 111 300 sentences as training data, 44 700 as validation data, and 3 000 sentences set aside as test data for evaluation. This dataset is available for download at the South African Centre for Digital Language Resources website.\footnote{Available online at: https://rma.nwu.ac.za/index.php/resource-catalogue/autshumato-english-setswana-multi-bilingual-corpus.html}

\subsection{Models}
\label{models}

Limited work has been done using NMT techniques for African languages. In fact, as far as the authors can tell, this is the first work using modern NMT techniques for translation for South African official languages. We thus selected two recent NMT architectures, convolutional sequence-to-sequence and Transformer, to compare to existing research. The existing research uses phrase-based SMT for English-to-Setswana translation.

The Fairseq(-py) tookit was used to model the convolutional sequence to sequence model \cite{gehring2017convolutional}. The model used was one of Fairseq's named architectures ``fconv''. The learning rate was set to 0.25, a dropout of 0.2, and the maximum tokens for each mini-batch was set to 4000. The dataset was preprocessed using Fairseq's preprocess script to build the vocabularies and to binarize the dataset. To decode the test data, beam search was used, with a beam width of 5.

The Tensor2Tensor implementation of Transformer was used \cite{tensor2tensor}. The model was trained for 125K steps . The learning rate was set to 0.4, with a batch size of 1024, and a learning rate warm-up of 45000 steps. The dataset was encoded using the Tensor2Tensor data generation algorithm which invertibly encodes a native string as a sequence of subtokens \cite{tensor2tensor}. Beam search was used to decode the test data, with a beam width of 4.

Training took less than 12 hours for both algorithms on a NVIDIA K80 GPU.

\section{Results}

Section~\ref{quant} describes the quantitative performance of the models by comparing BLEU scores, while a brief qualitative analysis is performed in in Section~\ref{qual}.

\subsection{Quantitative}
\label{quant}

\begin{table}[t]
  \caption{BLEU scores of the trained models on the Autshumato dataset.}
  \label{result-table}
  \centering
  \begin{tabular}{ll}
    \toprule

    Model    & BLEU \\
    \midrule
    From \cite{wilken2012developing}  & 28.8  \\
    ConvS2S & 27.77\\
    Transformer (uncased) & 33.53 \\
    Transformer (cased) & 33.12 \\
    \bottomrule
  \end{tabular}
\end{table}

According to the BLEU scores reported in Table~\ref{result-table}, ConvS2S achieved a BLEU score of 27.77, which is 1.02 BLEU points below the performance of the phrase-based English-Setswana system from \cite{wilken2012developing}. The Transformer model significantly outperformed the phrase-based English-Setswana system, by 5.33 BLEU points. Despite the fact neither ConvS2S nor Transformer had information from additional language models or linguistic pre-processing, the architectures performed extremely competitively, with Transformer achieving a new state-of-the-art of 33.12 for English-to-Setswana translation.

\subsection{Qualitative}
\label{qual}

\begin{table}[t]
  \caption{For each source sentence we show the reference translation, and the translations by the various models. We also show the translation of the results back to English, performed by a Setswana speaker.}
  \label{sent-result-table}
  \centering
  \begin{tabular}{p{3cm}|p{10cm}}
    \toprule
    Source  &  Take a shower rather than a bath as showers use less water and electricity.\\
    Reference & Dirisa šawara boemong jwa go tsena mo bateng ka gonne dišawara di dirisa eneji e nnye.\\
    ConvS2S & Tsaya tsia go na le kgonagalo ya gore go na le metsi a a ka nnang dirisang metsi le motlakase.\\
    Setswana-to-Eng Speaker ConvS2S & Take caution of the possibility that there is water that can remain use water and electricity. \\
    Transformer &  Tsamaya matsidinyana ka bonako go na le dibata ka di dirisa metsi a a kwa tlase le motlakase.\\
    Setswana-to-Eng Speaker Transformer &  Go really fast instead of baths because they use minimal water and electricity.\\\hline
    Source  &  This is to protect the abuse of children and young workers.\\ 
    Reference & Ntlha e e botlhokwa go sireletsa tiriso ya bana le bašwa.\\
    ConvS2S & Se ke go sireletsa tshotlako ya bana le badiri ba bannye. \\
    Setswana-to-Eng Speaker ConvS2S &  This is to protect the abuse of children and young workers. \\
    Transformer & Se ke go sireletsa tshotlako ya bana le bašwa.  \\
    Setswana-to-Eng Speaker Transformer & This is to protect the abuse of children and adolescents.\\ \hline
    Source  &  An example is a rural job creation project that aimed to deal with the fact that 40\% of people in a community are unemployed.   \\ 
    Reference & Sekao ke porojeke ya go tlholwa ga ditiro kwa magaeng go samagana le ntlha ya gore 40\% ya batho mo setšhabeng ga ba dire. \\
    ConvS2S &  Sekao ke porojeke ya go tlhola ditiro tsa go tlhola ditiro tse di ikaelelang go samagana le ntlha e e 40\% ya batho mo baaging ba sa dire.\\
    Setswana-to-Eng Speaker ConvS2S &  Example is a project that will create jobs that will create jobs that aim at dealing with the point that that 40\% of people in the society don't work. \\
    Transformer & Sekao ke porojeke ya go tlhama ditiro tsa magae e e ikaelelang go samagana le ntlha ya gore 40\% ya batho ba ba sa direng. \\
    Setswana-to-Eng Speaker Transformer & Example is a rural job creation project that aims to deal with the issue that 40\% of people are not working. \\
    \bottomrule
  \end{tabular}
\end{table}

Table~\ref{sent-result-table} shows qualitative results for specific sentences from our test set. In order to understand the feasibility of using such models, a Setswana speaker translated the English-to-Setswana translations generated by our model back to English. Although not perfect, the translations capture much of the meaning from the original sentence. Impressively, the translations also use synonyms for other concepts: for example, Transformer translated "40\% of people in a community are unemployed" to the Setswana equivalent of "40\% of people are not working".

In conjunction with the quantitative results, these results confirm our hypothesis that the use of NMT systems, in particular the Transformer model, can improve the state of the art in English to Setswana translation.

\section{Conclusion}

Due to the rising need for African translations of online educational resources, the development of accurate machine translation systems for low-resourced languages has become an issue of importance.

We showed that state-of-the-art NMT architectures can significantly outperform existing SMT architectures for translation from English to Setswana with minimal hyperparameter optimization, and only a small amount of training time. This result suggests the promise of using the Transformer architecture to train models to translate other African languages. Future work includes training the Transformer architecture on multiple African languages at once, and then specialising on a specific language, as is done for Romanian by Gu \textit{et al} \cite{gu2018meta}.

The source code and the data used are available at https://github.com/LauraMartinus/ukuxhumana.

\section{Acknowledgements}
We would like to thank the organisers of the Deep Learning Indaba. Without the Indaba we would never have met, nor would we have had the resources and confidence to pursue and submit such research. Thank you to Guy Bosa for aiding us with our qualitative translations.

\bibliographystyle{unsrt}
\bibliography{main}

\begin{thebibliography}{10}

\bibitem{spaull2013south}
Nicholas Spaull.
\newblock South africa’s education crisis: The quality of education in south
  africa 1994-2011.
\newblock {\em Johannesburg: Centre for Development and Enterprise}, pages
  1--65, 2013.

\bibitem{parikh2002utilizing}
Mihir Parikh and Sameer Verma.
\newblock Utilizing internet technologies to support learning: an empirical
  analysis.
\newblock {\em International Journal of Information Management}, 22(1):27--46,
  2002.

\bibitem{w3tech2018usage}
W3Techs.com.
\newblock Usage of content languages for websites.
\newblock \url{https://w3techs.com/technologies/overview/content_language/all},
  2018.

\bibitem{stats2012census}
SA~Stats.
\newblock Census 2011 statistical release.
\newblock {\em Pretoria, South Africa: Statistics South Africa}, 2012.

\bibitem{chepkemoi_2017}
Joyce Chepkemoi.
\newblock What languages are spoken in rwanda?
\newblock
  \url{https://www.worldatlas.com/articles/what-languages-are-spoken-in-rwanda.html},
  Jul 2017.

\bibitem{nationmaster}
Ethiopia language stats.
\newblock
  \url{http://www.nationmaster.com/country-info/profiles/Ethiopia/Language},
  2018.

\bibitem{muaka2011language}
Leonard Muaka.
\newblock Language perceptions and identity among kenyan speakers.
\newblock In {\em Selected Proceeding of the 40th Annual Conference on African
  Linguistics}, pages 217--230. Cascadilla Proceedings Project Somerville, MA,
  2011.

\bibitem{peter2016rwth}
Jan-Thorsten Peter, Tamer Alkhouli, Andreas Guta, and Hermann Ney.
\newblock The rwth aachen university english-romanian machine translation
  system for wmt 2016.
\newblock In {\em Proceedings of the First Conference on Machine Translation:
  Volume 2, Shared Task Papers}, volume~2, pages 356--361, 2016.

\bibitem{gu2018universal}
Jiatao Gu, Hany Hassan, Jacob Devlin, and Victor~OK Li.
\newblock Universal neural machine translation for extremely low resource
  languages.
\newblock {\em arXiv preprint arXiv:1802.05368}, 2018.

\bibitem{ronald2006statistical}
Kato Ronald and Etienne Barnard.
\newblock Statistical translation with scarce resources: a south african case
  study.
\newblock 2006.

\bibitem{wilken2012developing}
Ilana Wilken, Marissa Griesel, and Cindy McKellar.
\newblock Developing and improving a statistical machine translation system for
  english to setswana: a linguistically-motivated approach.
\newblock In {\em Twenty-Third Annual Symposium of the Pattern Recognition
  Association of South Africa}, page 114, 2012.

\bibitem{gehring2017convolutional}
Jonas Gehring, Michael Auli, David Grangier, Denis Yarats, and Yann~N Dauphin.
\newblock Convolutional sequence to sequence learning.
\newblock {\em arXiv preprint arXiv:1705.03122}, 2017.

\bibitem{gu2018meta}
Jiatao Gu, Yong Wang, Yun Chen, Kyunghyun Cho, and Victor~OK Li.
\newblock Meta-learning for low-resource neural machine translation.
\newblock {\em arXiv preprint arXiv:1808.08437}, 2018.

\bibitem{cmckellar2018autshumato}
Cindy McKellar.
\newblock Autshumato english-setswana parallel corpora.
\newblock
  \url{https://rma.nwu.ac.za/index.php/resource-catalogue/autshumato-english-setswana-multi-bilingual-corpus.html},
  2018.

\bibitem{tensor2tensor}
Ashish Vaswani, Samy Bengio, Eugene Brevdo, Francois Chollet, Aidan~N. Gomez,
  Stephan Gouws, Llion Jones, \L{}ukasz Kaiser, Nal Kalchbrenner, Niki Parmar,
  Ryan Sepassi, Noam Shazeer, and Jakob Uszkoreit.
\newblock Tensor2tensor for neural machine translation.
\newblock {\em CoRR}, abs/1803.07416, 2018.

\end{thebibliography}

% \section*{References}

% References follow the acknowledgments. Use unnumbered first-level
% heading for the references. Any choice of citation style is acceptable
% as long as you are consistent. It is permissible to reduce the font
% size to \verb+small+ (9 point) when listing the references. {\bf
%   Remember that you can use more than eight pages as long as the
%   additional pages contain \emph{only} cited references.}
% \medskip

% \small

% [1] Alexander, J.A.\ \& Mozer, M.C.\ (1995) Template-based algorithms
% for connectionist rule extraction. In G.\ Tesauro, D.S.\ Touretzky and
% T.K.\ Leen (eds.), {\it Advances in Neural Information Processing
%   Systems 7}, pp.\ 609--616. Cambridge, MA: MIT Press.

% [2] Bower, J.M.\ \& Beeman, D.\ (1995) {\it The Book of GENESIS:
%   Exploring Realistic Neural Models with the GEneral NEural SImulation
%   System.}  New York: TELOS/Springer--Verlag.

% [3] Hasselmo, M.E., Schnell, E.\ \& Barkai, E.\ (1995) Dynamics of
% learning and recall at excitatory recurrent synapses and cholinergic
% modulation in rat hippocampal region CA3. {\it Journal of
%   Neuroscience} {\bf 15}(7):5249-5262.

\end{document}